\documentclass[11pt, a4paper, logo, one column]{craftjarvis}

\pdftrailerid{redacted}

\makeatletter
\renewcommand\bibentry[1]{\nocite{#1}{\frenchspacing\@nameuse{BR@r@#1\@extra@b@citeb}}}
\makeatother

\usepackage{xcolor}
\definecolor{Gray}{gray}{0.9}
\definecolor{mygreen}{rgb}{0.0, 0.5, 0.0}
\definecolor{myred}{rgb}{0.8, 0.25, 0.33}
\definecolor{myblue}{rgb}{0.19, 0.55, 0.91}
\definecolor{uclablue}{rgb}{0.15, 0.45, 0.68}
\definecolor{zihaoblue}{rgb}{0.25, 0.44, 0.88}
\definecolor{zihaored}{rgb}{0.79, 0.36, 0.27}
\definecolor{boxgreen}{rgb}{0.02, 0.66, 0.02}
\definecolor{boxred}{rgb}{0.66, 0.1, 0.1}
\definecolor{boxblue}{rgb}{0.01, 0.01, 0.73}
\definecolor{langyellow}{RGB}{248, 211, 119}
\definecolor{visgreen}{RGB}{159, 206, 99}
\definecolor{actblue}{RGB}{194, 214, 236}
\definecolor{craftjarviscolor}{RGB}{100, 72, 247}

\definecolor{mygray}{gray}{0.4}

\usepackage{times}
\usepackage{graphicx}
\usepackage{amssymb}
\usepackage{url}

\usepackage{ulem}
\usepackage{marvosym}

\usepackage{microtype}
\usepackage{subfigure}
\usepackage{booktabs}

\usepackage{amsmath}
\usepackage{mathtools}
\usepackage{amsthm}

\usepackage{listings}
\usepackage{etoc}
\usepackage{algorithm}
\usepackage{algorithmic}
\usepackage{lipsum}
\usepackage{pifont}
\usepackage{wrapfig}
\usepackage{colortbl}
\usepackage{acronym}
\usepackage{multicol}
\usepackage{multirow}
\usepackage{makecell}
\usepackage{enumitem}
\usepackage{tabularx}
\usepackage{colortbl}
\usepackage{array}
\usepackage{xspace}
\usepackage[skip=3pt,font=small]{caption}
\usepackage{xspace}
\usepackage{mdframed}

\usepackage[export]{adjustbox}

\usepackage[utf8]{inputenc} % allow utf-8 input
\usepackage[T1]{fontenc}    % use 8-bit T1 fonts
\usepackage{hyperref}       % hyperlinks
\usepackage{amsfonts}       % blackboard math symbols
\usepackage{nicefrac}       % compact symbols for 1/2, etc.
\usepackage{mathtools}
\usepackage{amssymb}
\usepackage{amsthm}

\renewcommand{\paragraph}[1]{\noindent\textbf{#1.}}
\makeatletter
\DeclareRobustCommand\onedot{\futurelet\@let@token\@onedot}
\def\@onedot{\ifx\@let@token.\else.\null\fi\xspace}

\makeatother

\usepackage{xcolor}

\usepackage[most]{tcolorbox}

\newtcolorbox[list inside=prompt,auto counter,number within=section]{prompt}[1][]{
    colbacktitle=black!60,
    coltitle=white,
    fontupper=\footnotesize,
    boxsep=5pt,
    left=0pt,
    right=0pt,
    top=0pt,
    bottom=0pt,
    boxrule=1pt,
    #1,
}

\acrodef{llms}[LLMs]{Large Language Models}
\acrodef{mlms}[MLMs]{Multimodal Language Models}
\acrodef{rag}[RAG]{Retrieval Augmented Generation}
\acrodef{cot}[CoT]{chain-of-thought}

\usepackage[authoryear, sort&compress, round]{natbib} 

\title{OpenHA: A Series of Open-Source Hierarchical Agentic Models in Minecraft}

\reportnumber{} % Leave blank if n/a 

\renewcommand{\today}{Sep 2025} 

\author[1]{Zihao~Wang$^\dagger$}
\author[1]{Muyao~Li$^\dagger$}
\author[1]{Kaichen~He$^\dagger$}
\author[1]{Xiangyu~Wang}
\author[1]{Zhancun~Mu}
\author[2]{Anji~Liu}
\author[1]{Yitao~Liang}

\correspondingauthor{Yitao Liang<yitaol@pku.edu.cn>. 
$\dagger$ indicates equal contribution.\\
Zihao Wang<zhwang@stu.pku.edu.cn>, Muyao Li<2200017405@stu.pku.edu.cn>, Kaichen He<hkc4623@gmail.com>, Xiangyu Wang<2300017816@stu.pku.edu.cn>, Zhancun Mu<muzhancun@stu.pku.edu.cn>, Anji Liu<anjiliu@comp.nus.edu.sg>
}

\affil[1]{Peking~University}
\affil[2]{National University of Singapore}
\affil[ \hspace{-0.73ex}]{All authors are affiliated with Team CraftJarvis} 

\begin{abstract}

The choice of action representation is a critical yet unresolved challenge in developing capable, end-to-end trainable agents. This paper first presents a large-scale, systematic comparison of prominent action abstractions for Vision-Language-Action (VLA) or hierarchical agent models in the open-ended Minecraft. Our analysis reveals that no single action space is universally optimal; instead, the most effective abstraction is highly task-dependent, creating a dilemma for building generalist agents.
To resolve this, we introduce Chain of Action (CoA), a novel framework that unifies high-level planning and low-level control within a single, monolithic VLA model. CoA treats an abstracted action not as a command for a separate policy, but as an intermediate reasoning step—akin to a chain of thought—that guides the generation of the final, executable action. Furthermore, we demonstrate that an All-in-One agent trained on a diverse mixture of action spaces using the CoA paradigm learns a more robust and generalizable policy. This unified agent achieves a new state-of-the-art, improving the overall task success rate over strong, specialized baselines.
To foster reproducible research, we release the OpenHA (Open Hierarchical Agents) suite, which includes our comprehensive benchmark of over 800 distinct tasks, curated datasets, source code, and all pretrained model checkpoints at: \url{https://github.com/CraftJarvis/OpenHA}.
\end{abstract}

\begin{document}

% \correspondingauthor{Xiaojian~Ma,~Yitao~Liang\\
% Zihao Wang<zhwang@stu.pku.edu.cn>, Shaofei Cai<caishaofei@stu.pku.edu.cn>, Anji Liu<liuanji@cs.ucla.edu>, Xiaojian~Ma<xiaojian.ma@ucla.edu>, Yitao Liang<yitaol@pku.edu.cn>}

\maketitle

\section{Introduction}\label{sec:intro}

Autonomous agents, in their early designs, were predominantly constructed as modular systems leveraging pre-trained models, including Large Language Models (LLMs)~\citep{llama,chatgpt,qwen3,claude4} and Vision-Language Models (VLMs)~\citep{gpt-4,qwen2vl,gemini,seed15vl}. These agents often integrated a static foundation model with external modules—such as memory, planning, and tool use—to facilitate interaction within an environment~\citep{weng2023agent,yang2023auto0gpt,wang2024rat,cheng2024exploring}. However, such modular pipelines cannot inherently undergo end-to-end training, resulting in a critical limitation: the agent’s performance is constrained by the capabilities of the pre-trained model~\citep{voyager,yu2025memagent}. More importantly, the agent cannot learn and improve from its interactions with the environment~\citep{proagent,li2025chain}.

To address this constraint, recent advancements have turned toward the development of end-to-end pre-trained agent models that learn directly from interaction data~\citep{uitars,operator}. These models embed the capacity for action generation within the agent itself. Within this framework, two main methodologies for modeling an agent’s actions have emerged. The first, Vision-Language-Action (VLA) models~\citep{openvla}, tokenize low-level actions—such as keyboard inputs or robotic movements—into sequences that are compatible with the vocabulary of language models. This enables the model to predict and generate low-level actions directly from the interaction context. However, a significant challenge arises from the semantic gap between the high-level abstractions expressed by language and the continuous, fine-grained nature of low-level actions.

To overcome this issue, a second class of agent models adopts a hierarchical architecture~\citep{rth,palme,gr00t}. In these models, a high-level language model predicts an abstracted action, which is then decoded into a sequence of low-level actions by a learned action tokenizer~\citep{omnijarvis,groot}. While this hierarchical approach separates reasoning and control, a key question arises: \textit{how can abstracted actions be effectively defined and formulated from the continuous stream of low-level experiences?}

Despite the widespread adoption of various action tokenizers~\citep{zhong2025survey}, a clear, standardized evaluation remains lacking. Most action tokenizers are tested on domain-specific benchmarks and proprietary datasets, making direct and fair comparisons difficult~\citep{rt-h,rttraj,openvla}. This lack of standardization obscures a crucial question: \textit{which action representation is the most effective?} To fill this gap, we introduce a unified benchmark for the open-ended environment of Minecraft~\citep{minedojo}, where we conduct a large-scale comparison of prominent abstracted action spaces across 1000 diverse tasks. Our empirical findings reveal that the effectiveness of an action tokenizer is highly task-dependent, with no single representation demonstrating universal superiority.

Due to the traditional abstracted action being used only in hierarchical agent architecture, we further explored whether these abstracted actions are also beneficial for the VLA model. We introduce Chain-of-action (CoA) framework, which integrates the strengths of both hierarchical agents and VLA models.
This observation motivates our primary contribution: the introduction of the Chain of Action (CoA) framework, which integrates the strengths of both hierarchical agents and VLA models. In CoA, action generation is structured as a two-step, auto-regressive process: first, the model generates a high-level abstracted action, which serves as an intermediate ``thinking'' or ``reasoning''; then, conditioned on this thought, the model generates the final low-level action. For instance, to execute the command ``chop a tree'', the CoA agent first predicts a grounding action such as \texttt{Approach(object=<tree>, coordinate=[136, 287])}, followed by the sequence of low-level actions required to reach the position and perform the action. This approach effectively synergizes high-level reasoning with low-level control.

In addition, inspired by the discovery that different abstracted actions excel in distinct task domains, we propose an All-in-One training strategy. The key insight is that, by grounding the model’s final output in low-level actions, CoA enables a unified prediction structure across all action spaces. This allows us to train a single agent on a composite dataset containing multiple action representations. The resulting agent can master a diverse range of actions, outperforming specialist models trained on a single type of action.

The main contributions of this work are fourfold:
\begin{itemize}
\item We provide a large-scale, systematic analysis demonstrating that the optimal abstracted action space is task-dependent, with different action representations excelling in distinct task domains.
\item We introduce the Chain of Action framework, which synergizes high-level reasoning and low-level control by using abstracted actions as intermediate plans. This novel approach bridges the gap between hierarchical agents and VLA models, offering superior performance compared to traditional dual-system architectures.
\item We demonstrate that training a single agent on mixed datasets of diverse action spaces enhances its ability to generalize across tasks, improving its overall decision-making capabilities.
\item We release a comprehensive suite of resources, including trained checkpoints, code, and datasets for various hierarchical agents, hierarchical VLA models and \textbf{OpenHA} model to support further research in action representation and generalization.
\end{itemize}

\section{Abstracted Actions in Hierarchical Agents}\label{sec:preliminary}

\subsection{Hierarchical Agent Structure}
Hierarchical agents model the agent's decision-making process through a hierarchical policy decomposition. This process consists of two levels: first, a high-level pretrained LLM-based agentic model generates an abstracted action based on the task instruction and the current observation; second, a low-level policy acts as the \textbf{action tokenizer} and uses this abstracted action as guidance to produce a primitive, executable action in the environment.
This two-level procedure can be formally expressed as:
\begin{align}
    A \sim \pi_{AR}(\cdot \mid obs, ins), a \sim \pi_{policy}(\cdot \mid obs, A) \label{eq:causal_action}
\end{align}
Here, $ins$ denotes the human-provided textual instruction, and $obs \in \mathbb{R}^{H\times W\times 3}$ represents the current visual observation.
The variable $A \in \mathcal{A}_{abs}$ signifies an \textbf{abstracted action} sampled from a high-level action space $\mathcal{A}_{abs}$, generated by the high-level policy $\pi_{AR}$. Subsequently, $a \in \mathcal{A}_{env}$ represents a \textbf{primitive environmental action} from the low-level, executable action space $\mathcal{A}_{env}$, generated by the low-level policy $\pi_{policy}$ as the action tokenizer~\citep{zhong2025survey}. In domains like computer interaction, these primitive actions correspond to raw inputs such as discrete keyboard presses or mouse movements.

The specific formulation of the abstracted action $A$ is a critical design choice and varies significantly across different agent architectures. It can range from symbolic, language-based skills (e.g., ``pick up the apple'')~\citep{palme}, to continuous, grounded interaction trajectories~\citep{rttraj}, textual descriptions of motion (e.g., ``go forward'')~\citep{rth}, or even discrete latent codes learned by encode-decoder models~\citep{omnijarvis,gr00t}.

In typical implementations, the high-level auto-regressive models $\pi_{AR}$ are instantiated as a large-scale Vision-Language Model (VLM)~\citep{qwen2vl,llama3,gpt4}, leveraging its powerful reasoning capabilities to process multimodal inputs and generate the abstracted actions $A$. In contrast, the low-level policy $\pi_{policy}$ is often a more lightweight, specialized model, such as a Transformer with less than 100M parameters~\citep{transformerxl,transformer}, tasked with translating the plan $A$ into a sequence of environment-specific primitive actions $a$~\citep{steve1,groot}. 

This hierarchical formulation is general enough to subsume many recent Vision-Language-Action (VLA) models~\citep{diffusion_policy,octo,rt-1,rt-2,tracevla, hiRT,dexgraspvla,roboVLA,upvla,objectvla,chatvla,combat_vla}. These non-hierarchical or "flat" VLA models can be viewed as a special case of our framework where the hierarchy collapses and the policy $\pi_\text{policy}$ is the language tokenizer itself~\citep{qwen}. In this scenario, the high-level policy is trained to directly output the specification of the primitive action, effectively making the abstracted action space equivalent to the primitive one ($A\equiv a$). Consequently, the low-level policy $\pi_{policy}$ reduces to a trivial identity function, as the planning and execution steps are merged into a single end-to-end network.

\subsection{Action Spaces in Hierarchical Agents}

\begin{table}[t]
\centering
\caption{Definitions and examples of the different action spaces in agents. 
% The full action spaces for different actions can be found in the Appendix. 
}
\label{tab:action_space}
\resizebox{\textwidth}{!}{%
\renewcommand\arraystretch{1.2}
\begin{tabular}{@{}llll@{}}
\toprule
Sysbol & Defination & Action Types & Example \\ \midrule
$a$ & Env Action & 24 & \begin{tabular}[c]{@{}l@{}}\{"camera": {[}-1, -9{]}, "ESC": 0, "back": 0, "drop": 0, "forward": 0, "hotbar.1-9": 0, "inventory": 0, "jump": 0, \\ "left": 0, "right": 0, "sneak": 0, "sprint": 0, "swapHands": 0, "attack": 0, "use": 0, "pickItem": 0\}\end{tabular} \\
$RA$ & Raw Action & 35 & "\textless{}|reserved\_token\_1|\textgreater \textless{}|reserved\_token\_7|\textgreater \textless{}|reserved\_token\_9|\textgreater \textless{}|reserved\_token\_2|\textgreater " \\
$TA$ & Text Action & 3 & \begin{tabular}[c]{@{}l@{}}"Action: keyDown(keys=(keyboard.left.control, keyboard.w, keyboard.a))"\\ "Action: move(dx=`4.0', dy=`-1.0') and keyDown(keys=(keyboard.left.control, keyboard.w))"\end{tabular} \\
$MA$ & Motion Action & - & "Action: Go forward, Turn left." \\
$GA$ & Grounding Action & 8 & \begin{tabular}[c]{@{}l@{}}"Action: Mine(object=`oak\_log', position={[}100, 200{]})"\\ "Action: Approach(object=`sheep', position={[}200, 300{]})"\end{tabular} \\
$LA$ & Latent Action & 15000 & "\textless{}|reserved\_token\_2|\textgreater{}" \\ \bottomrule
\end{tabular}%
}
\end{table}

The efficacy of a hierarchical agent is intrinsically tied to the design of its action space, A. This space defines the vocabulary of behaviors the agent can command~\citep{zhong2025survey}. In this work, we investigate a spectrum of action spaces, ranging from low-level environmental controls to various high-level abstractions. A summary and examples of these spaces are provided in Table~\ref{tab:action_space}.

\paragraph{Primitive Environmental Actions ($a$)} 
At the most fundamental level is the primitive action space dictated by the environment simulator. In our context, this corresponds to low-level keyboard and mouse operations. To be processed by a language-based model, these raw actions must be tokenized. We explore two common strategies: 1) \textbf{Reserved Tokens as Raw Action ($RA$)}: This method involves discretizing the continuous action space and mapping each discrete action combination to a unique, reserved token within the model's vocabulary~\citep{rt2,openvla}. 2) \textbf{Textual Serialization as Text Action $TA$)}: Alternatively, actions can be serialized into descriptive strings (e.g., \texttt{move(dx=4, dy=-1)}). This approach leverages the model's existing text-processing capabilities without requiring new vocabulary tokens.

\paragraph{Language Skills $SA$} A natural choice for high-level models based on VLMs or LLMs is to use natural language skills as the abstracted action space. These skills typically describe a complete, goal-oriented behavior, often with a verb-object structure like ``chop down trees'' or ``open the chest''. Executing such a skill requires a sophisticated, language-conditioned low-level policy that interprets the command and, conditioned on the current visual observation, decodes it into a sequence of primitive actions to achieve the goal~\citep{deps,palme}.

\paragraph{Motion Primitives $MA$} Occupying an intermediate level of abstraction, motion primitives represent temporally extended but object-agnostic movements. For instance, a continuous sequence of ``\texttt{press W and Left.Shift}'' can be abstracted into a single ``sprint forward'' command. Similarly, a series of mouse movements can be categorized as ``turn left'' in an egocentric frame. These actions are more abstract than primitive environmental actions but are lower-level than language skills, as they do not specify an object of interaction~\citep{steve1,rth}.

\paragraph{Grounding Actions $GA$} While powerful, pure language skills face scalability challenges, as the low-level policy must learn to interpret every possible object for a given skill. Grounding actions address this by creating a structured, multimodal representation. They retain the symbolic verb from language skills (e.g., \texttt{mine}) but replace the object's name with its grounded spatial coordinates in the visual observation (e.g., \texttt{coordinate=[100, 200]}). This decouples the ``what'' from the ``where,'' enabling a single, grounding-conditioned policy to execute a skill on any object by simply targeting its location, significantly improving generalization~\citep{dexgraspvla,rocket1,lee2025molmoact}.

\paragraph{Latent Actions $LA$} The aforementioned action spaces rely on human-defined heuristics and feature engineering. In contrast, latent actions are learned directly from data in a self-supervised manner. This is typically achieved using an encoder-decoder framework, where an encoder compresses sequences of primitive actions into a continuous latent embedding. To make these embeddings compatible with auto-regressive generation, a Vector-Quantized Variational Autoencoder (VQ-VAE) is often employed to discretize the continuous latent space into a finite codebook of tokens~\citep{vq}. The high-level policy then generates these learned latent tokens as its abstracted actions~\citep{deng2025open,yuan2024pre,BehaviorTransformer}.

 \begin{figure*}[t!]
    \centering
    \includegraphics[width=0.95\linewidth]{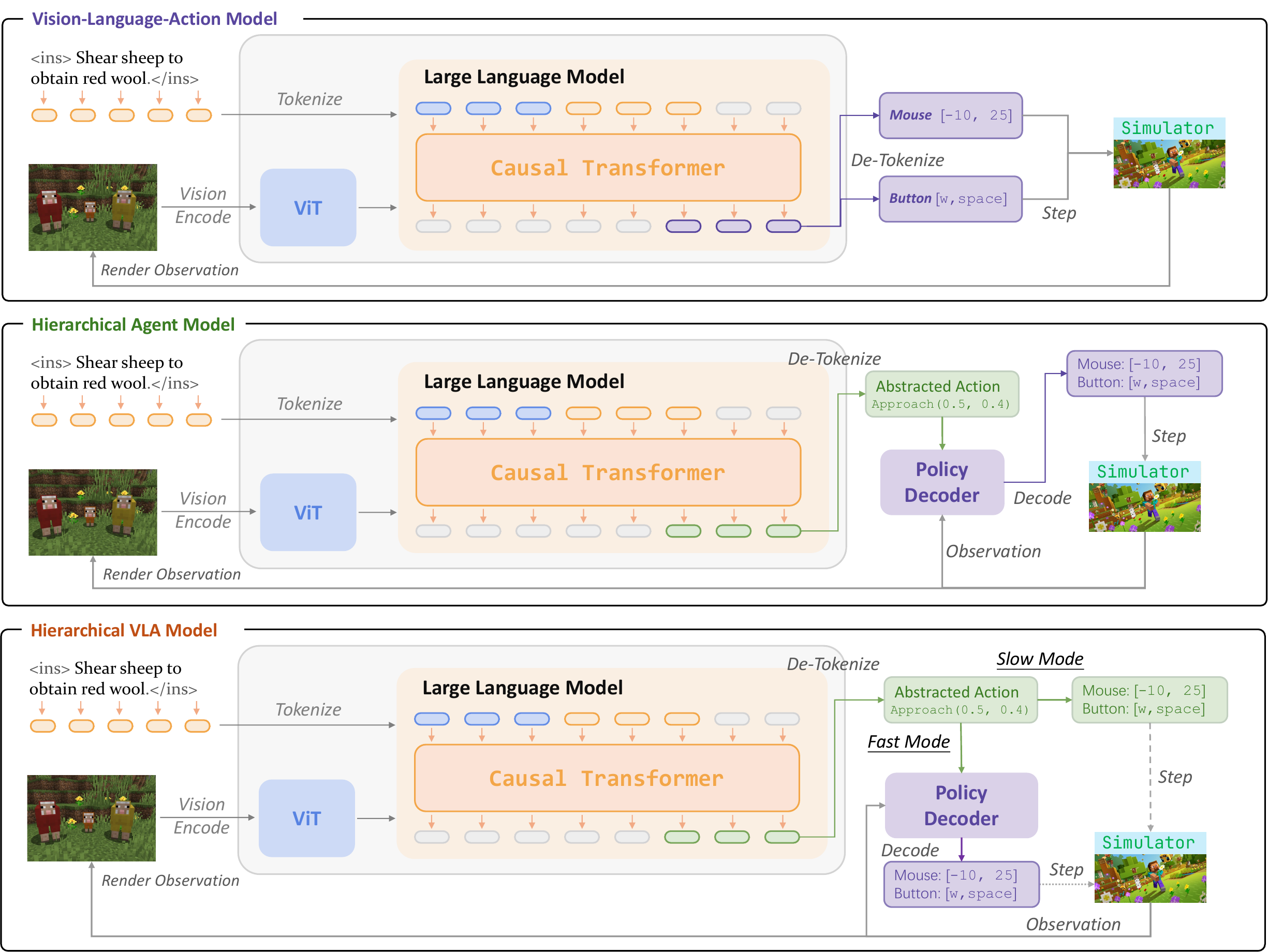}
    % \vspace{-0.6em}
    \caption{
    Comparisons between end-to-end vision-language-action (VLA) models and hierarchical agent (HA) models. 
    The core difference between HA and the VLA model is that HA uses a policy-based action tokenizer to predict actions, while the vision-language-action model uses a text tokenizer to decode actions. We further propose integrating the abstracted actions of the hierarchical agent into the training and inference processes of VLA, resulting in the Hierarchical VLA model, which has both fast and slow inference modes, using a policy-based action tokenizer and a text tokenizer to decode environmental actions, respectively. 
    }
    \label{fig:vla_vs_ha}
    % \vspace{-0.3 in}
    % \vspace{-1.5em}
\end{figure*}

\section{Synergizing abstracted and low-level actions within VLA models}\label{sec:method}

\subsection{Taking Abstracted Actions as Thoughts for VLA models}

While abstracted actions simplify the decision-making landscape for VLM-based agents, their conventional implementation within a rigid two-stage, hierarchical architecture introduces significant drawbacks~\citep{jarvis1,palme}. Typically, this separation prevents end-to-end training, requiring the low-level policy $\pi_{\text{policy}}$ to be trained independently. This not only complicates the training pipeline but also establishes the low-level policy as a potential performance bottleneck, as the agent's overall capability is capped by its decoder's effectiveness.

To circumvent these limitations, we propose a method to integrate the benefits of action abstraction directly into a unified, end-to-end trainable Vision-Language-Action (VLA) model. We introduce the \textbf{Chain of Action (CoA)}, an auto-regressive formulation that reframes action generation as a single, sequential process. The core idea is to treat the high-level abstracted action $A$ not as a command for a separate module, but as an intermediate "thought" or reasoning step that guides the subsequent prediction of the low-level environmental action $a$.

This approach leverages the powerful sequence modeling capabilities of modern auto-regressive transformers~\citep{embodiedcot,chainofthought,transformer}. 
Instead of predicting the final action $a$ directly from the observation $obs$ and instruction $ins$, the model first generates the abstracted action $A$ and then, conditioned on this self-generated context, predicts the primitive action $a$. We formalize this by modeling the joint probability of the action chain $(A, a)$. 
Given an instruction $\text{ins}$ and an observation $\text{obs}$, the likelihood of generating the sequence is factorized using the chain rule of probability:
\begin{equation}
    P(A, a \mid \text{ins}, \text{obs}) = \underbrace{P(a \mid \text{ins}, \text{obs}, A)}_{\text{Action Generation}} \cdot \underbrace{P(A \mid \text{ins}, \text{obs})}_{\text{"Thought" Generation}}
    \label{eq:coa}
\end{equation}
Here, a single auto-regressive policy, $\pi_{\theta}$, models the entire generation process. 
The term $P(A \mid \text{ins}, \text{obs})$ represents the model generating its intermediate thought (the abstracted action). 
The term $P(a \mid \text{ins}, \text{obs}, A)$ represents the model generating the final executable action, crucially conditioned on its own preceding thought. 
The entire model $\pi_{\theta}$ is trained end-to-end by maximizing the log-likelihood of this joint probability over a dataset of expert trajectories.

To illustrate, consider the task "\textsc{chop down a tree}." 
A standard VLA model must implicitly reason about the tree's location, its distance, and the required navigation, all while directly outputting a complex sequence of low-level actions (e.g., {\footnotesize\texttt{press W}}, {\footnotesize\texttt{turn left}}, {\footnotesize\texttt{look up}}, {\footnotesize\texttt{left click mouse}}). 
This implicit reasoning is difficult to learn. 
In contrast, the CoA agent explicitly decomposes this problem:
\begin{enumerate}
    \item \textbf{Thought Generation}: The model first generates a grounded action as a thought, such as {\footnotesize\texttt{$GA$: Attack(object=oak\_log, coordinate=[100, 200])}}.
    \item \textbf{Action Generation}: This generated text now becomes part of the model's context. Conditioned on this explicit target, the subsequent prediction of low-level actions becomes a simpler, more grounded task. The model can observe that the target at \texttt{[100, 200]} is far away and to the left, making the generation of {\footnotesize\texttt{a: press W}} and {\footnotesize\texttt{$a$: turn left}} a direct consequence of the explicit reasoning step.
\end{enumerate}
In this way, the CoA formulation grounds complex reasoning in an explicit, intermediate representation, breaking down a difficult prediction problem into a more manageable, sequential one and seamlessly unifying hierarchical reasoning within a single end-to-end model.

\subsection{Hierarchical VLA Models with Fast or Slow Inference Methods}

The Chain of Action (CoA) formulation is not only a training paradigm but also an architectural principle that affords significant flexibility at inference time. As illustrated in Figure~\ref{fig:vla_vs_ha}, our model can operate in two distinct inference modes, allowing for a dynamic trade-off between reasoning depth and computational efficiency.

\paragraph{Decoupled Inference Mode (Fast)}
This mode emulates the classic hierarchical agent architecture. The primary large language model, which we refer to as the high-level policy $\pi_{AR}$, is queried only to generate the abstracted action $A$. The subsequent task of decoding this plan into a sequence of primitive actions $\{a^1, \ldots, a^k\}$ is offloaded to a separate, lightweight low-level policy $\pi_{policy}$ as de-tokenizer. This approach significantly reduces the inference load on the large VLM, as it is invoked only once per high-level abstracted action, while the more efficient $\pi_{policy}$ can interact with the environment for multiple steps. However, this mode has two primary drawbacks. First, it typically requires a multi-stage training process: the low-level policy $\pi_{policy}$ must be trained first to serve as a decoder, followed by the training of the high-level policy $\pi_{AR}$. Second, the overall performance is potentially bottlenecked by the capability of the smaller, specialized low-level policy.

\paragraph{Unified Autoregressive Mode (Slow)}
In contrast, this mode fully leverages the end-to-end nature of the CoA framework. The single, unified model $\pi_\theta$ is responsible for generating the entire action sequence autoregressively, first producing the abstracted action $A$ as a plan, and then immediately generating the corresponding low-level action $a$, as described in Equation~\ref{eq:coa}. This approach ensures that the full reasoning capacity of the large model is applied to both high-level planning and low-level execution, eliminating any potential performance bottlenecks from a separate decoder. The primary trade-off is the increased computational cost at inference time, as the VLM must generate a longer sequence of tokens for each environmental action, resulting in slower decision-making.

A crucial advantage of our framework is that these two inference modes do not require separately trained models. A single agent can be endowed with both capabilities through a unified, end-to-end training process on a mixed dataset. During training, we provide the model with examples formatted for both objectives. The final output format is controlled via a specific system prompt provided as input to the model. This allows the model to learn both to generate high-level plans alone and to follow through with low-level execution within a single set of weights.

This methodology effectively merges the traditional hierarchical agent structure and the end-to-end VLA paradigm into a single, versatile architecture. It empowers a practitioner to dynamically select the most suitable inference mode at runtime—choosing the fast, decoupled mode for simpler tasks or resource-constrained environments, and the slower, more powerful unified mode for tasks demanding complex reasoning.

\subsection{OpenHA: Training One VLA Model to Master Diverse Abstracted Action Spaces}

As we establish in our experimental analysis (Section \ref{sec:experiments}), the optimal choice of abstracted action space is not universal but is instead highly contingent on the specific task. This finding presents a practical dilemma: must we develop and maintain a suite of specialist agents, each tailored to a different action representation? Or, more ambitiously, can a single, universal agent learn to master a diverse repertoire of action abstractions simultaneously?

To answer this question, we propose the \textbf{All-in-One} training strategy, which aims to produce a single VLA model proficient in multiple action spaces. A naive approach would be to simply train a model on a heterogeneous dataset of trajectories, $\mathcal{D} = \bigcup_k \mathcal{D}_k$, where each sub-dataset $\mathcal{D}_k$ contains trajectories annotated with a different high-level action space $\mathcal{A}^{(k)}$ (e.g., Grounding Actions, Motion Actions, etc.). However, such a strategy faces a critical challenge: the different action spaces are fundamentally disparate. Without a shared basis, the model may treat them as isolated prediction domains, failing to generalize or transfer knowledge between them.

Our Chain of Action (CoA) framework provides a principled solution to this challenge. By adopting the CoA data format for all trajectories—$(ins, \{(o_t, A_t, a_t)\}_{t=1}^T)$—we ensure that every high-level abstracted action $a$, regardless of its origin space, is ultimately grounded in the same primitive environmental action space $\mathcal{A}$. The primitive action $a_t$ acts as a \textbf{common currency}, creating a shared semantic ground that connects all high-level abstractions.

This common grounding enables the model to learn the underlying relationships and functional equivalences between different action representations. For example, the model can infer that a high-level plan expressed as a Grounding Action like {\small\texttt{$GA$: Approach(object=sheep)}} and one expressed as a Motion Action like {\small\texttt{$MA$: Go forward}} can be contextually similar if they both result in the same primitive action sequence {\small\texttt{$a$: press(w)}}. The training objective remains identical to that defined in Equation~\ref{eq:coa}, maximized over the full, heterogeneous dataset $\mathcal{D}$.

We hypothesize that this All-in-One training strategy will yield two significant benefits. First, it will produce a highly versatile agent capable of understanding and generating multiple types of action abstractions. Second, and more importantly, by learning a more holistic and interconnected representation of behavior, the All-in-One agent will develop a more robust and generalizable internal policy, ultimately outperforming any specialist agent trained on only a single form of action abstraction.

\section{Experiments and Insights}
\label{sec:experiments}

In this section, we present a comprehensive suite of experiments designed to empirically validate our proposed frameworks and investigate key questions surrounding action representation for open-ended agents. We begin by detailing our experimental setup, including the training protocols for our models, the large-scale benchmark constructed for evaluation, and the metrics used to measure performance. Our analysis is then structured to systematically answer three central research questions:

\begin{enumerate}
    \item Within a traditional Hierarchical Agent (HA) paradigm, what type of abstracted action space is most effective in a complex, open-ended environment like Minecraft?
    \item Can incorporating these abstractions as intermediate reasoning steps within a unified VLA model improve its overall performance?
    \item Does training a single agent on a heterogeneous mixture of action spaces enable it to learn a more generalizable and effective decision-making policy?
\end{enumerate}

\subsection{Training Details}

\paragraph{Training Models} To ensure a fair and controlled comparison across all experiments, we initialize all agents from a common foundation model. Specifically, we use the \textbf{Qwen2-VL-7B}~\citep{qwen2vl} model that has been post-trained on a large corpus of Minecraft-specific VQA and captioning data, as detailed in~\citep{jarvisvla}. For our baseline VLA implementation, we adopt the model configuration from JARVIS-VLA. For agents that utilize a latent action space, we follow the action tokenizer configuration of OmniJARVIS~\citep{omnijarvis}.

\paragraph{Dataset Curation.} Our raw training data is derived from the OpenAI Video Pre-Training dataset~\citep{vpt}, which contains a large volume of expert trajectories in Minecraft. However, the original VPT dataset only provides pairs of visual observations and corresponding low-level environmental actions $(o_t, a_t)$. To facilitate the training of agents with high-level action spaces, we developed a \textbf{programmatic labeling pipeline}. This pipeline applies a set of rule-based heuristics to convert the raw, low-level action sequences into their corresponding high-level abstracted action representations ($A_t$), including Motion, Grounding, and Language Skill actions.
This process yields three distinct types of trajectory datasets used in our experiments:
\begin{itemize}
    \item \textbf{High-Level Action Data ($D_A$):} Trajectories containing only observations and their corresponding high-level abstracted actions, formatted as $\{(o_t, A_t)\}_{t=1}^T$.
    \item \textbf{Low-Level Action Data ($D_a$):} The original trajectories containing only observations and primitive environmental actions, formatted as $\{(o_t, a_t)\}_{t=1}^T$.
    \item \textbf{Chain-of-Action Data ($D_\text{CoA}$):} Trajectories containing the complete triplet of observation, the generated high-level action, and the corresponding primitive action, formatted as $\{(o_t, A_t, a_t)\}_{t=1}^T$.
\end{itemize}

\begin{table}[]
\centering
    \caption{Training recipes for different agent models.}
    \label{tab:training_recipes}
    \resizebox{0.9\textwidth}{!}{%
    \renewcommand\arraystretch{1.2}
\begin{tabular}{@{}lcccccccccc@{}}
\toprule
Model &  & HA &  & VLA &  & \multicolumn{2}{c}{Hierarchical VLA} &  & \multicolumn{2}{c}{OpenHA Model} \\ 
\cmidrule(r){1-1} \cmidrule(lr){3-3} \cmidrule(lr){5-5} \cmidrule(lr){7-8} \cmidrule(l){10-11}
Stage &  & 1 &  & 1 &  & 1 & 2 &  & 1 & 2 \\
Datasets &  & $D^A[GA|MA|SA]$ &  & $D^a$ &  & $D^A{+}D^a$ & $D^\text{CoA}$ &  & $D^A[MA|GA]{+}D^a$ & $D^\text{CoA}[MA+GA]$ \\
Consumed Tokens &  & 1.58B &  & 3.59B &  & 3.33B & 0.27B &  & 3.40B & 0.22B \\
Learning Rate &  & 5e-6 &  & 1e-5 &  & 1e-5 & 5e-6 &  & 1e-5 & 5e-6 \\
Training Steps &  & 4000 &  & 10400 &  & 10800 & 400 &  & 8200 & 300 \\
\bottomrule
\end{tabular}}
\end{table}

\paragraph{Training Recipes}
We employ distinct training recipes tailored to the specific architecture of each agent category to ensure a fair evaluation of their respective paradigms.
Standard end-to-end VLA models are trained exclusively on the low-level action dataset, $D_a$. The training objective is to maximize the likelihood of the primitive action given the current observation, $P(a_t | o_t, ins)$.
For the Hierarchical Agent (HA) baselines, we focus on training the high-level VLM policy. These models are trained exclusively on the high-level action dataset, $D_A$. Their objective is to learn the policy $P(A_t | o_t, ins)$. The low-level policies used to decode $A_t$ are trained separately according to their original implementations.
The chain-of-action based VLA agents, which unify abstracted action $A_t$ and low-level action $a_t$ in a single model, are trained using a two-stage curriculum designed to first build foundational knowledge and then learn the crucial link between abstracted actions and environmental actions.
The model is first trained on a mixture of the high-level and low-level datasets, $D_A \cup D_a$. This stage serves as a multi-task pre-training phase, familiarizing the model with the distinct vocabularies and formats of both abstracted plans and primitive actions. The pre-trained model is then fine-tuned exclusively on the Chain-of-Action dataset, $D_{CoA}$. This stage is critical for teaching the model to bridge planning and execution by learning the joint probability $P(A_t, a_t | o_t, ins)$, as defined in Equation~\ref{eq:coa}.
For the All-in-One OpenHA agent, the datasets $D_A$ and $D_{CoA}$ used in this curriculum are composed of a heterogeneous mixture of all considered action abstraction types. We use the TRL~\citep{trl} and VeOmni~\citep{veomni} library to train the agent models and use the vLLM library to support efficient agent inference~\citep{vllm}.
The detailed training parameters and datasets are listed in the Table~\ref{tab:training_recipes}.

\subsection{Experimental Setup}

\paragraph{Simulator and Environment} We employ Minecraft (Version 1.16.5) as our primary testbed~\citep{minerl}. The agent's observation space consists solely of first-person RGB visual frames at a resolution of $360 \times 640 \times 3$. The action space comprises discretized, human-like mouse and keyboard controls, including continuous mouse displacement, mouse clicks, and various keyboard inputs. 
% Further details on the observation and action space specifications are provided in Appendix~\ref{app:observation_and_action}.

\paragraph{Evaluation Benchmarks}
Previous work on Minecraft agents has often been evaluated on a limited number of canonical tasks~\citep{vpt,steve1,groot}. 
Although MCU~\citep{mcu} and Minedojo~\citep{minedojo} both claim to have around 3,000 tasks, these are usually automatically generated by LLMs, and cannot guarantee a fair and stable success rate for the agents.
To enable a more comprehensive and fair comparison of different action spaces, we construct a large-scale benchmark comprising nearly 1,000 distinct tasks. All tasks are designed and verified by hand. 
We categorize these tasks into three groups based on the primary capabilities required: 
\begin{itemize}
    \item \textbf{Embodied} tasks requiring navigation and physical interaction within the 3D world (e.g., finding and chopping down a specific type of tree).
    \item \textbf{GUI} tasks that involve complex interactions with graphical user interfaces, such as crafting items at a crafting table or smelting ores in a furnace.
    \item \textbf{Combat} tasks that require the agent to engage with hostile mobs, focusing on survival and combat skills.
\end{itemize}
To rigorously test for generalization, we ensure that all evaluation environments are out-of-distribution with respect to the training data. This is achieved by sampling novel initial world seeds and spawn locations for each task evaluation from Minecraft's vast procedural generation space.

\paragraph{Compared Models and Baselines}
Our primary analysis focuses on a controlled, systematic comparison of agents equipped with the different action spaces detailed in Section \ref{sec:method}, including our proposed CoA-based hierarchical VLA models and OpenHA models. To contextualize our results within the broader literature, we also compare against established, specialized policies with fewer than 1B parameters that were trained on the VPT dataset. These baselines include the original \textbf{VPT}~\citep{vpt}, \textbf{ROCKET-1}~\citep{rocket1}, and \textbf{STEVE-1}~\citep{steve1}.

\paragraph{Metrics and Evaluation Protocol}
For a comprehensive assessment, each agent is evaluated on every task in our benchmark for a minimum of three independent runs, each with a different world seed. To obtain more statistically robust results on a representative sample, we selected 10 tasks of varying difficulty (easy, medium, and hard) from each of the three groups for a more intensive evaluation of over 10 runs each. Our primary metrics are:
\begin{enumerate}
    \item \textbf{Success Rate:} The percentage of runs in which the agent successfully completes the given task.
    \item \textbf{Steps to Completion:} The number of environmental steps required to complete a task.
\end{enumerate}
For our final analysis, we report the average Success Rate and average number of steps, aggregated across all tasks within each of the three benchmark groups.

% \subsection{Main Results}

\begin{table}[]
    \centering
    \caption{
    Evaluation results of Minecraft agents on over 800+ tasks.
    For each task category, we report the average steps to finish tasks, the average success rates across mini sets, and the Average Success Rate across all tasks in that category (\textit{ASR} $\pm$ standard deviation). 
    Results highlighted in {\color{blue}{blue}} represent the second-best performances, while those in {\color{red}{red}} indicate the state-of-the-art performance for each metric across all agents. `-' signifies tasks where the agent failed to achieve any success.
    }
    \label{tab:q1}
    \resizebox{\textwidth}{!}{
    \renewcommand\arraystretch{1.2}
    \begin{tabular}{@{}lllccccccccccc@{}}
\toprule
 &  &  & \multicolumn{3}{c}{Embodied Tasks} &  & \multicolumn{3}{c}{GUI Tasks} &  & \multicolumn{3}{c}{Combat Tasks} \\ \cmidrule(lr){4-6} \cmidrule(lr){8-10} \cmidrule(l){12-14} 
\multirow{-2}{*}{Model} & \multirow{-2}{*}{\begin{tabular}[c]{@{}l@{}}Action\\ Tokenizer\end{tabular}} & \multirow{-2}{*}{Inference} & Steps & ASR (Mini) & ASR (All) &  & Steps & ASR (Mini) & ASR (All) &  & Steps & ASR (Mini) & ASR (All) \\ \midrule
\multicolumn{14}{l}{\cellcolor[HTML]{ECF4FF}{ \textit{\textbf{Previous Methods}}}} \\
VPT & - & End-to-end & 377 & 10.1$^{\pm3.6}$ & 6.0$^{\pm11.4}$ &  & 398 & 0.7$^{\pm0.1}$ & 0.8$^{\pm3.3}$ &  & 396 & 3.6$^{\pm7.7}$ & 3.6$^{\pm7.7}$ \\
STEVE-1 & - & End-to-end & 384 & 8.4$^{\pm3.0}$ & 8.0$^{\pm17.0}$ &  & 391 & 0.0 & 3.2$^{\pm8.4}$ &  & 395 & 4.9$^{\pm1.8}$ & 3.9$^{\pm12.0}$ \\
ROCKET-1 & - & Workflow & 392 & 19.2$^{\pm6.1}$ & 18.9$^{\pm24.3}$ &  & N/A & 0.0 & 0.0 &  & 320 & \color{blue}{ \textbf{29.8$^{\pm9.0}$}} & \color{blue}{ \textbf{27.9$^{\pm29.3}$}} \\
JARVIS-VLA & - & End-to-end & 305 & 31.0$^{\pm12.7}$ & 30.0$^{\pm35.4}$ &  & 339 & \color{blue}{ \textbf{25.3$^{\pm5.7}$}} & \color{blue}{ \textbf{25.1$^{\pm23.9}$}} &  & 352 & 18.3$^{\pm5.2}$ & 18.5$^{\pm22.7}$ \\ \midrule
\multicolumn{14}{l}{\cellcolor[HTML]{F0FBEF}{ \textit{\textbf{VLM-based Hierarchical Agents}}}} \\
LatentHA & Latent & Hierarchical & 363 & 27.3$^{\pm37.4}$ & 24.4$^{\pm31.1}$ &  & 393 & 3.5$^{\pm8.7}$ & 3.0$^{\pm7.5}$ &  & 371 & 8.2$^{\pm15.6}$ & 8.5$^{\pm17.9}$ \\
MotionHA & Motion & Hierarchical & 336 & 31.6$^{\pm10.1}$ & 27.4$^{\pm35.2}$ &  & N/A & 0.0 & 0.0 &  & 392 & 9.1$^{\pm3.9}$ & 4.3$^{\pm10.8}$ \\
GroundingHA & Grounding & Hierarchical & \color{blue}{ \textbf{290}} &\color{red}{ \textbf{39.7$^{\pm13.7}$}} & \color{red}{\textbf{37.1$^{\pm38.5}$}} &  & 380 & 3.7$^{\pm2.3}$ & 6.7$^{\pm10.8}$ &  & 346 & 28.2$^{\pm6.2}$ & 26.5$^{\pm23.4}$ \\
SkillHA & Skill & Hierarchical & 365 & 13.8$^{\pm7.7}$ & 11.3$^{\pm14.5}$ &  & 397 & 3.4$^{\pm0.8}$ & 6.3$^{\pm9.2}$ &  & 393 & 3.4$^{\pm0.8}$ & 6.5$^{\pm9.3}$ \\ \midrule
\multicolumn{14}{l}{\cellcolor[HTML]{FEF1F1}{ \textit{\textbf{Vision-Language-Action Models}}}} \\
TextVLA & Text & End-to-end & 321 & 23.9$^{\pm8.9}$ & 27.0$^{\pm17.0}$ &  & \color{red}{\textbf{291}} & 14.0$^{\pm4.1}$ & 25.8$^{\pm14.3}$ &  & \color{blue}{ \textbf{317}} & 27.1$^{\pm11.8}$ & 10.0$^{\pm6.1}$ \\
OpenHA & Text & End-to-end & \color{red}{\textbf{287}} & \color{blue}{ \textbf{37.0$^{\pm15.9}$}} & \color{blue}{ \textbf{30.1$^{\pm13.9}$}} &  & \color{blue}{ \textbf{314}} & \color{red}{\textbf{33.3$^{\pm13.3}$}} & \color{red}{\textbf{32.5$^{\pm9.2}$}} &  & \color{red}{\textbf{316}} & \color{red}{\textbf{40.0 $^{\pm19.6}$}} & \color{red}{\textbf{31.9$^{\pm13.7}$}}\\ \bottomrule
\end{tabular}
}
% \vspace{-0.2 in}
\end{table}

\subsection{Fair Evaluation of Action Spaces in Hierarchical Agents}
\label{sec:fair_evaluation_ha}

In this section, we seek to answer a fundamental question: \textbf{within a traditional Hierarchical Agent (HA) paradigm, what type of abstracted action space is most effective in a complex, open-ended environment like Minecraft?} To this end, we conducted a broad evaluation across our three distinct task categories. To ensure a fair and controlled comparison, each VLM-based HA was trained on a dataset containing an equivalent number of action tokens from its respective action space.

First, we re-evaluated previously open-sourced methods on our new, large-scale benchmark. As shown in the top section of Table~\ref{tab:q1}, these methods exhibit a significant drop in performance compared to their originally reported scores. For example, STEVE-1 achieves only 8.0\% Average Success Rate (ASR) on Embodied tasks. We attribute this to the fact that prior work was often trained and evaluated on a narrower set of tasks. While effective on canonical benchmarks like acquiring wool from sheep, these agents lack the generalizability to succeed on more diverse challenges, such as fishing or defeating creepers, resulting in near-zero success rates on many tasks.

In contrast, our VLM-based Hierarchical Agents demonstrate substantially stronger performance, underscoring the potential of leveraging large, pre-trained models for agentic control. However, our central finding is that \textbf{no single action space is universally optimal; instead, performance is highly contingent on the task domain}, as illustrated in Figure~\ref{fig:vla_vs_ha} and detailed in Table~\ref{tab:q1}.

The agent utilizing grounding actions, \textbf{GroundingHA}, emerges as the most capable model within the HA category. It achieves the highest ASR among HAs in both Embodied ($37.1\%$) and Combat ($26.5\%$) tasks. Conversely, the agent using object-agnostic motion primitives, \textbf{MotionHA}, shows respectable performance in navigation-heavy Embodied tasks ($27.4\%$) but fails completely in GUI tasks ($0.0\%$), which require precise, object-centric interactions. The other agents, \textbf{LatentHA} and \textbf{SkillHA}, show moderate but not leading performance in any category.
We further compare the performance and extend to each individual task. We find that the best action tokenizer under each task in hierarchical agents is not uniform. For example, in the ``kill the sheep'' task, \textbf{MotionHA} showed a nearly 100\% success rate, making it the best model; while in the ``Chop down trees'' task, \textbf{SkillHA} achieved a 95\% success rate, making it the best model.
The root cause of these performance disparities lies in the inherent expressive capabilities and limitations of each action space.
\begin{itemize}
    \item \textbf{Grounding Actions (GroundingHA)} are powerful because they explicitly bind an action (e.g., `\texttt{mine}') to a specific entity's pixel coordinates. In addition, this type of action representation also utilizes the visual grounding ability of the foundational visual language model and its rich world knowledge. This is highly effective when the target is visible, explaining its strong performance in Combat and object-interaction embodied tasks. Its primary theoretical weakness arises in partially observable scenarios where the target is off-screen. In such cases, the agent might resort to a generic `\texttt{Explore}' action, which is less efficient than a directed `\texttt{turn left}' command from \textbf{MotionHA}. Nonetheless, its strong empirical results suggest this is often a successful strategy.
    \item \textbf{Motion Actions (MotionHA)} excel at expressing navigational intent without needing a visible target. However, this object-agnostic nature is also a critical flaw, as the agent cannot specify \textit{what} to interact with, rendering it incapable of performing GUI-based crafting or targeted actions.
    \item \textbf{GUI Task Performance.} For GUI interactions, precise grounding is non-negotiable. While \textbf{GroundingHA}'s ASR of $6.7\%$ on these complex tasks is modest, it still outperforms \textbf{MotionHA}, highlighting the importance of object-centric representation.
\end{itemize}

Ultimately, this analysis reveals that the choice of action tokenizer for a hierarchical agent involves a critical trade-off between navigational generality and object-centric precision. This lack of a single ``\textbf{best}'' action space motivates our core proposal: synergizing these different representations within a single, more capable model.

\begin{table}[]
    \centering
    \caption{
    Ablation study on the trade-off between inference mode and performance for agents using Motion and Grounding actions. We compare traditional Hierarchical Agents (e.g., \textbf{MotionHA}) using a \textbf{Fast}, decoupled inference mode against our unified models (e.g., \textbf{MotionVLA}) trained with Chain of Action (CoA) and using the \textbf{Slow}, unified autoregressive mode. FPS denotes inference speed (Frames Per Second). The CoA models consistently achieve higher Success Rates (SR) at the cost of lower FPS.
    }
    \label{tab:q2}
    \resizebox{\textwidth}{!}{
    \renewcommand\arraystretch{1.2}
    \begin{tabular}{@{}lllccccccccccc@{}}
\toprule
& \multicolumn{1}{l}{} &  & \multicolumn{3}{c}{Embodied Tasks} &  & \multicolumn{3}{c}{GUI Tasks} &  & \multicolumn{3}{c}{Combat Tasks} \\ 
\cmidrule(lr){4-6} \cmidrule(lr){8-10} \cmidrule(l){12-14} 
\multirow{-2}{*}{Model} & \multicolumn{1}{l}{\multirow{-2}{*}{Inference}} & \multirow{-2}{*}{FPS} 
& Steps & SR (Mini) & SR (All) &  & Steps & SR (Mini) & SR (All) &  & Steps & SR (Mini) & SR (All) \\ 
\midrule
\multicolumn{14}{l}{\cellcolor[HTML]{FEFFF0}{ \textit{Baseline}}} \\
TextVLA & - & 1.36 & 321 & 23.9$^{\pm8.9}$ & 27.0$^{\pm17.0}$ &  & 291 & 14.0$^{\pm4.1}$ & 25.8$^{\pm14.3}$ &  & 317 & 27.1$^{\pm11.8}$ & 10.0$^{\pm6.1}$ \\ 
\midrule
\multicolumn{14}{l}{\cellcolor[HTML]{F0F4FF}{ \textit{Motion Actions}}} \\
MotionHA & Fast & 3.75 & 336 & 31.6$^{\pm10.1}$ & 27.4$^{\pm35.2}$ &  & N/A & 0.0$^{\pm0.0}$ & 0.0$^{\pm0.0}$ &  & 392 & 9.1$^{\pm3.9}$ & 4.3$^{\pm10.8}$ \\
MotionVLA & Slow & 0.98 & 330 & 29.7$^{\pm12.2}$ & 24.8$^{\pm10.9}$ &  & -- & -- & -- &  & 344 & 26.3$^{\pm13.1}$ & 25.6$^{\pm9.8}$ \\ 
\midrule
\multicolumn{14}{l}{\cellcolor[HTML]{F0FBEF}{ \textit{Grounding Actions}}} \\
GroundingHA & Fast & 5.61 & 290 & 39.7$^{\pm13.7}$ & 37.1$^{\pm38.5}$ &  & 380 & 3.7$^{\pm2.3}$ & 6.7$^{\pm10.8}$ &  & 346 & 28.2$^{\pm6.2}$ & 26.5$^{\pm23.4}$ \\
GroundingVLA & Slow & 1.22 & 316 & 35.5$^{\pm11.2}$ & 30.1$^{\pm13.9}$ &  & -- & -- & -- &  & 342 & 30.6$^{\pm15.6}$ & 27.4$^{\pm15.9}$ \\ 
\bottomrule
\end{tabular}
}
\end{table}

\subsection{Experiments on Chain-of-Action Training}
\label{sec:exp_coa}

Having established that different action abstractions are specialized for different tasks, we now investigate our central hypothesis: \textbf{can incorporating these abstractions as intermediate reasoning steps within a unified VLA model improve its overall performance?}

To do this, we compare our unified Chain of Action (CoA) models against their traditional hierarchical counterparts. As detailed in Section 3.2, our architecture supports both a \textbf{Fast (Decoupled)} mode, which mimics a traditional HA by using a separate lightweight decoder, and a \textbf{Slow (CoA)} mode, which performs full autoregressive generation of the plan and action. This allows for a direct comparison of the two paradigms.

The results, presented in Table~\ref{tab:q2}, reveal a distinct and consistent trade-off between inference efficiency and task performance. 
For agents using Motion Actions, the traditional hierarchical model (\textbf{MotionHA}) is highly efficient, operating at 3.75 FPS. Our unified model using the CoA paradigm (\textbf{MotionVLA}) is computationally more intensive, running at 0.98 FPS. However, this additional computational cost at inference time effectively compensates for MotionHA's weakness in Combat tasks: the ASR increases nearly sixfold, from 4.3\% to 25.6\%. 
A similar trend can be observed for Grounding Actions. The success rate in Combat tasks improves from 26.5\% to 27.4\%.
% For agents using Grounding Actions, the traditional hierarchical model (\textbf{GroundingHA}) is highly efficient, operating at 5.61 FPS. Our unified model using the CoA paradigm (\textbf{GroundingVLA}) is computationally more intensive, running at 1.22 FPS. However, this increased computational investment at inference time yields a dramatic improvement in capability: the ASR on Embodied tasks rises from 21.7\% to 37.1\%, and the Combat ASR nearly doubles from 14.8\% to 26.5\%. A similar pattern holds for agents using Motion Actions, where the slower \textbf{MotionVLA} significantly outperforms its faster counterpart, \textbf{MotionHA}, particularly in Combat tasks (25.6\% vs 4.3\% ASR).

Crucially, the performance gains are not merely relative to the HA baselines. Our strongest CoA model, \textbf{GroundingVLA}, also substantially outperforms the robust \textbf{TextVLA} baseline in both Embodied (30.1\% vs. 27.0\% ASR) and Combat (27.4\% vs. 10.0\% ASR) domains. This provides a clear, affirmative answer to our initial question: explicitly generating an abstracted action as an intermediate ``thought'' allows the VLA model to better structure its decision-making process, breaking down complex problems into a more manageable sequence of planning and execution.

We posit that this demonstrates a novel and powerful form of \textbf{inference-time scaling} for autonomous agents. While performance is commonly scaled by increasing model size or training data (at training time), CoA offers a complementary axis for improvement: allocating more computational resources at inference time to a structured, explicit reasoning process. By ``thinking'' more before acting, the agent achieves a higher level of performance, confirming that abstracted actions, when integrated thoughtfully, significantly improve the capabilities of VLA models.

\begin{table}[]
    \centering
    \caption{
    Performance comparison between specialist agents (trained on a single action space) and our universal \textbf{OpenHA} agent (trained on all action spaces). For each comparison, OpenHA is prompted to use the same inference format as the corresponding specialist (e.g., MotionCoA). The results show that the universally-trained OpenHA model consistently outperforms the specialists, demonstrating positive knowledge transfer across action spaces. 
    }
    \label{tab:q3}
    \resizebox{\textwidth}{!}{
    \renewcommand\arraystretch{1.2}
    \begin{tabular}{@{}llccccccccccc@{}}
\toprule
 &  & \multicolumn{3}{c}{Embodied Tasks} &  & \multicolumn{3}{c}{GUI Tasks} &  & \multicolumn{3}{c}{Combat Tasks} \\ \cmidrule(lr){3-5} \cmidrule(lr){7-9} \cmidrule(l){11-13} 
\multirow{-2}{*}{Model} & \multirow{-2}{*}{Inference} & Steps & SR (Mini) & SR (All) &  & Steps & SR (Mini) & SR (All) &  & Steps & SR (Mini) & SR (All) \\ \midrule
\multicolumn{13}{l}{\cellcolor[HTML]{FEFFF0}{ \textit{Text Action}}} \\
TextVLA & Text & 321 & 23.9$^{\pm8.9}$ & 27.0$^{\pm17.0}$ &  & 291 & 14.0$^{\pm4.1}$ & 25.8$^{\pm14.3}$ &  & 317 & 27.1$^{\pm11.8}$ & 10.0$^{\pm6.1}$ \\
OpenHA & Text & 326 & 37.0$^{\pm15.9}$ & 27.5$^{\pm13.8}$ &  & 314 & 33.3 $^{\pm13.3}$ & 32.5$^{\pm9.2}$ &  & 304 & 40.0$^{\pm19.6}$ & 31.9$^{\pm13.7}$ \\ \midrule
\multicolumn{13}{l}{\cellcolor[HTML]{F0F4FF}{ \textit{Motion Actions}}} \\
MotionVLA & MotionCoA & 317 & 28.9$^{\pm11.7}$ & 31.2$^{\pm14.1}$ &  &  -- & -- & -- &  & 355 & 19.2$^{\pm3.7}$ & 15.8$^{\pm4.2}$ \\
OpenHA & MotionCoA & 330 & 29.7$^{\pm12.2}$ & 24.8$^{\pm10.9}$ &  & 308 & 28.7$^{\pm10.0}$ & 29.5$^{\pm8.7}$ &  & 344 & 26.3$^{\pm13.1}$ & 25.6$^{\pm9.8}$ \\ \midrule
\multicolumn{13}{l}{\cellcolor[HTML]{F0FBEF}{ \textit{Grounding Actions}}} \\
GroundingVLA & GroundingCoA & 303 & 34.9$^{\pm13.2}$ & 24.9$^{\pm9.6}$ &  & -- & -- & -- & & 327 & 18.3$^{\pm3.6}$ & 14.8$^{\pm3.2}$ \\
OpenHA & GroundingCoA & 316 & 35.5$^{\pm11.2}$ & 30.1$^{\pm13.9}$ &  & 288 & 34.6$^{\pm6.6}$ & 35.1$^{\pm10.4}$ &  & 342 & 30.6$^{\pm15.6}$ & 29.8$^{\pm12.6}$ \\ \bottomrule
\end{tabular}
}
% \vspace{-0.2 in}
\end{table}

\subsection{Effectiveness of Training on Diverse Action Spaces}

In this section, we investigate the central hypothesis of our All-in-One approach: \textbf{does training a single agent on a heterogeneous mixture of action spaces enable it to learn a more generalizable and effective decision-making policy?} To answer this, we conduct a series of experiments comparing our universally-trained agent, \textbf{OpenHA}, against the specialist models.

To ensure a fair comparison, we evaluate OpenHA by constraining its output to match the format of each specialist agent. For example, when comparing against \textbf{MotionVLA}, we prompt OpenHA to generate actions using the \texttt{MotionCoA} format. This allows us to isolate the effect of the training data diversity. The results of this comparison are presented in Table~\ref{tab:q3}.

The data reveals a clear and consistent trend: \textbf{the universally-trained OpenHA agent almost universally outperforms the specialist models, even when using their native action format.} For example, when using the \texttt{GroundingCoA} inference format on Embodied tasks, OpenHA achieves an ASR of 30.1\% compared to the specialist \textbf{GroundingVLA}'s 24.9\%. This consistent improvement across different inference formats and task categories strongly suggests that the model benefits from the diverse training data, learning a more robust and comprehensive internal policy.
It shows that OpenHA has learned a generalized concept of object interaction from its exposure to Text and Grounding actions, and it can leverage this latent knowledge even when forced to express its intent through an object-agnostic action format. This confirms that the All-in-One training fosters a deeper understanding of tasks that transcends any single action representation.

These findings have significant implications for the future of general-purpose agents. They suggest that co-training on diverse action spaces is not merely a method for creating a versatile generalist agent but a powerful mechanism for improving an agent's core reasoning and generalization capabilities. This points toward a promising direction where a single, highly capable agent could be trained to operate across disparate domains—such as web browser~\citep{wei2025browsecomp}, mobile GUI~\citep{operator,uitars15}, MCP APIs~\citep{mcp,feng2025retool}, and code environments~\citep{yang2024swe,jimenez2023swe}—by unifying their respective action spaces into one cohesive model, a path explored in recent works on cross-platform agents~\citep{uitars2,mobileagentv3}.

\section{Conclusions}\label{sec:conclusions}

In this work, we first conduct a large-scale analysis demonstrating that the optimal abstracted action for autonomous agents is highly task-dependent. To address this, we introduce the \textbf{Chain of Action} framework, which unifies high-level planning and low-level control within a single end-to-end VLA model. Building on this, we show that an \textbf{All-in-One} agent, trained on a diverse mixture of action spaces, learns a more generalizable policy and achieves superior performance compared to any specialist agent. Our findings suggest that synergizing multiple action representations, rather than selecting a single one, is a crucial step towards developing more capable and general-purpose autonomous agents.

% \newpage

\section*{Acknowledgement}

We thank Haowei Lin, Shaofei Cai, Guangyu Zhao, Minghao Liu, and Xiaojian Ma for discussions. And we thank PsiBot Inc. and Jianxin Du for providing computing devices and infrastructure support.

% \section*{Limitations}

% ...

% \section*{Ethics Statement}

% ...

% Bibliography components
\bibliographystyle{abbrvnat}
\nobibliography*
\bibliography{main}

\newpage

\appendix
\renewcommand\thefigure{\thesection.\arabic{figure}}
\setcounter{figure}{0}
% \section{Technical Appendices and Supplementary Material}
% Technical appendices with additional results, figures, graphs and proofs may be submitted with the paper submission before the full submission deadline (see above), or as a separate PDF in the ZIP file below before the supplementary material deadline. There is no page limit for the technical appendices.

\section{Open-source Models and Datasets}

\subsection{Open-source Models}

\begin{table}[H]
\resizebox{\textwidth}{!}{%
\renewcommand\arraystretch{1.2}
\begin{tabular}{@{}lll@{}}
\toprule
Model Name & Params & Huggingface URL \\ \midrule
Minecraft-MotionHA-Qwen2VL-2509 & 7B & \url{https://huggingface.co/CraftJarvis/minecraft-motionha-qwen2vl-7b-2509} \\
Minecraft-PointHA-Qwen2VL-2509 & 7B & \url{https://huggingface.co/CraftJarvis/minecraft-pointha-qwen2vl-7b-2509} \\
Minecraft-TextVLA-Qwen2VL-2509 & 7B & \url{https://huggingface.co/CraftJarvis/minecraft-textvla-qwen2vl-7b-2509} \\
Minecraft-OpenHA-Qwen2VL-2509-Base & 7B & \url{https://huggingface.co/CraftJarvis/minecraft-openha-qwen2vl-7b-2509} \\ \bottomrule
\end{tabular}}
\end{table}

\subsection{Open-source Datasets}

\begin{table}[H]
\resizebox{\textwidth}{!}{%
\renewcommand\arraystretch{1.2}
\begin{tabular}{@{}lll@{}}
\toprule
Action Space & Dataset Size & Huggingface URL \\ \midrule
Motion Action & 1B Tokens & \url{https://huggingface.co/CraftJarvis/minecraft-motion-action-dataset} \\
Grounding Action & 1B Tokens & \url{https://huggingface.co/CraftJarvis/minecraft-grounding-action-dataset} \\
Text Action & 2B Tokens & \url{https://huggingface.co/CraftJarvis/minecraft-text-action-dataset} \\
Motion CoA & 0.5B Tokens & \url{https://huggingface.co/CraftJarvis/minecraft-motion-coa-dataset} \\
Grounding CoA & 0.5B Tokens & \url{https://huggingface.co/CraftJarvis/minecraft-grounding-coa-dataset} \\ \bottomrule
\end{tabular}}
\end{table}

\clearpage

\end{document}